\documentclass[letterpaper]{article} 
\usepackage{arXiv_aaai24}  
\usepackage{times}  
\usepackage{helvet}  
\usepackage{courier}  
\usepackage[hyphens]{url}  
\usepackage{graphicx} 
\usepackage{amsmath}
\usepackage{natbib}  
\usepackage{caption} 
\usepackage{amsfonts}
\usepackage{amssymb}
\usepackage{algorithm}
\usepackage{algorithmic}

\usepackage{newfloat}
\usepackage{listings}
\DeclareCaptionStyle{ruled}{labelfont=normalfont,labelsep=colon,strut=off} 
\floatstyle{ruled}
\newfloat{listing}{tb}{lst}{}
\floatname{listing}{Listing}
\title{Uncertainty Quantification for Forward and Inverse Problems of PDEs \\via Latent Global Evolution}
\author{
    Tailin Wu\textsuperscript{\rm 1}\equalcontrib,
    Willie Neiswanger\textsuperscript{\rm 2}\equalcontrib,
    Hongtao Zheng\textsuperscript{\rm 1}\equalcontrib,\\
    Stefano Ermon\textsuperscript{\rm 3},
    Jure Leskovec\textsuperscript{\rm 3}
}
\affiliations{
    \textsuperscript{\rm 1} School of Engineering, Westlake University\\
    \textsuperscript{\rm 2} Department of Computer Science, University of Southern California\\
    \textsuperscript{\rm 3} Computer Science Department, Stanford University

    \{wutailin, zhenghongtao\}@westlake.edu.cn, neiswang@usc.edu, \{ermon, jure\}@cs.stanford.edu\\
}

\usepackage{bibentry}

\begin{document}

\maketitle

\begin{abstract}
Deep learning-based surrogate models have demonstrated remarkable advantages over classical solvers in terms of speed, often achieving speedups of 10 to 1000 times over traditional partial differential equation (PDE) solvers. However, a significant challenge hindering their widespread adoption in both scientific and industrial domains is the lack of understanding about their prediction uncertainties, particularly in scenarios that involve critical decision making. To address this limitation, we propose a method that integrates efficient and precise uncertainty quantification into a deep learning-based surrogate model. Our method, termed Latent Evolution of PDEs with Uncertainty Quantification (LE-PDE-UQ), endows deep learning-based surrogate models with robust and efficient uncertainty quantification capabilities for both forward and inverse problems. LE-PDE-UQ leverages latent vectors within a latent space to evolve both the system's state and its corresponding uncertainty estimation. The latent vectors are decoded to provide predictions for the system's state as well as estimates of its uncertainty. In extensive experiments, we demonstrate the accurate uncertainty quantification performance of our approach, surpassing that of strong baselines including deep ensembles, Bayesian neural network layers, and dropout. Our method excels at propagating uncertainty over extended auto-regressive rollouts, making it suitable for scenarios involving long-term predictions. Our code is available at: 
\url{https://github.com/AI4Science-WestlakeU/le-pde-uq}. 
\end{abstract}

\section{Introduction}
\begin{figure*}[t]
\centering
\includegraphics[width=0.8\textwidth]{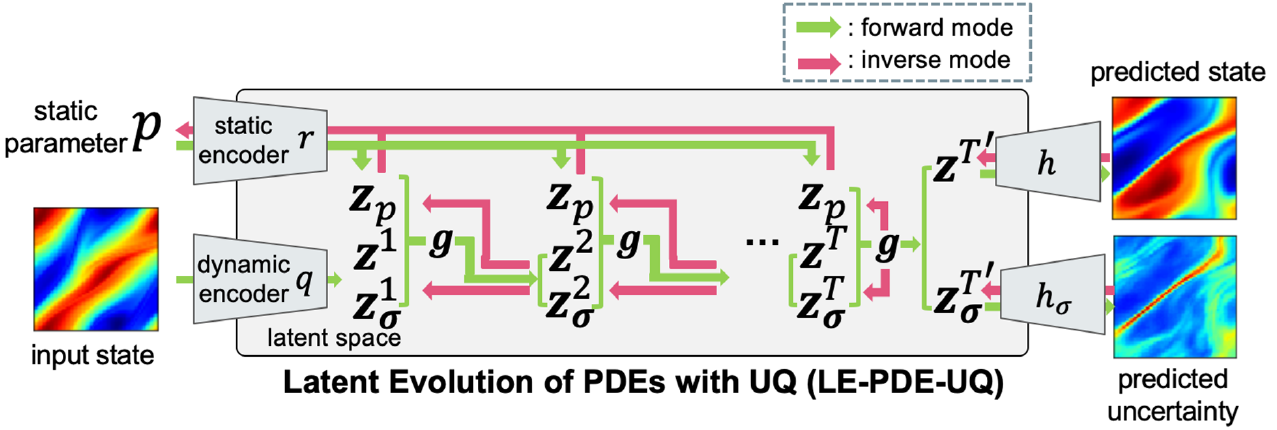} 
\caption{Schematic representation of the LE-PDE-UQ framework. In the forward mode(green), LE-PDE-UQ evolves the dynamics in the global latent space. In the inverse optimization mode (red), it optimizes the parameter p(e.g., the boundary) by unrolling the latent vectors. The compressed latent vectors and dynamics can significantly speed up both modes. The latent evolution model \textit{g} deterministically predicts a global latent vector \textit{z} encoding the state and a global latent vector $Z_{\sigma}$ encoding the uncertainty. On demand, they are decoded into the predicted state and the predicted uncertainty, respectively.}
\label{fig1}
\end{figure*}

Partial differential equations have wide-ranging applications in both scientific and engineering domains. It is noteworthy that time-dependent partial differential equations characterize the evolution of complex system states over time, serving as crucial tools for forward prediction and reverse optimization across various disciplines. These applications span a wide spectrum, including weather forecasting \cite{lynch2008origins,bi2023accurate}, nuclear fusion \cite{carpanese2021development}, jet engine design \cite{sircombe2006kinetic}, astronomical simulation \cite{courant1967partial}, molecular modeling \cite{lelievre2016partial}, 
and physical simulation modeling \cite{wu2022learningb,wu2023learning}, to name just a few. When addressing real-world challenges in science and engineering, the sheer volume of cells per time step can readily escalate into the millions or beyond. This complexity poses a substantial hurdle for conventional PDE solvers to ensure rapid solutions. In addition, inverse optimization such as inverse inference of system parameters also face similar scale challenges, besides the modeling of forward evolution \cite{biegler2003large}. Consequently, numerous deep learning-based alternative models have emerged that can accelerate the speed of partial differential equation solving by orders of magnitude (typically 10 to 1000 times), such as \cite{li2020fourier}.

However, recent neural network-based PDE solvers share a common drawback – they typically fail to provide any form of uncertainty estimation for their proposed solutions. This can lead to an overconfidence or underconfidence in the accuracy of the approximate solutions generated by PDE solvers, potentially resulting in relying on inaccurate approximations without any indication of associated risks. The concept of uncertainty quantification refers to the process of assessing and measuring the uncertainty associated with outcomes during prediction or optimization. In various scientific and engineering applications, the predictive results of models can be influenced by multiple factors, such as data noise, model uncertainty, parameter estimation, and more. Uncertainty quantification aims to provide information about the credibility or confidence level of the predictive outcomes, enhancing the understanding of the model's reliability. This information is valuable for considering uncertainty during decision-making processes. By quantifying uncertainty, we can better assess the stability, accuracy, and reliability of the model's predictions across different scenarios, enabling more informed decision-making.

Currently, most of the work related to uncertainty quantification in the process of PDE solving largely excludes consideration of temporal states \cite{winovich2019convpde,zhu2019physics}. This is mainly because, for time-domain partial differential equations, surrogate models need to perform autoregressive rollouts, which result in the accumulation of uncertainty over time. In the Julia programming language, there are libraries for uncertainty quantification (UQ), but none is designed specifically for neural surrogate models dealing with time-varying PDEs. An important consideration in temporal PDE uncertainty quantification is that the prediction process concurrent with PDE solving may adversely affect the speed and accuracy of the solution. Addressing this issue, we draw inspiration from LE-PDE (\cite{wu2022learninga}), which leverages latent representations to efficiently transfer valid information and capture the global features of the input state, reducing information redundancy and noise. The use of latent representations also significantly reduces the data dimensionality, thereby accelerating model inference and backpropagation.

To bridge the technical gap in uncertainty quantification for time-domain PDEs and address the aforementioned challenges, we introduce a novel framework named Latent Evolution of PDEs with Uncertainty Quantification (LE-PDE-UQ). This approach is simple, fast, and scalable, accurately quantifying the uncertainty arising in both the forward evolution and inverse optimization of PDEs. The comprehensive network structure of LE-PDE-UQ is showcased in Fig. \ref{fig1}. The method evolves the state of the system and the uncertainty estimation of the state by corresponding latent vectors in the latent space, and decodes them as state prediction and uncertainty estimation, respectively. The specific framework of LE-PDE-UQ will be described in Section 3 of this paper.

We have also shown in subsequent experiments that our method achieves state-of-the-art results in uncertainty estimation for forward evolution and inverse optimization of PDEs, is able to propagate uncertainty in long-term autoregressive prediction, outperforming strong baseline methods (e.g., deep ensembles, Bayes layers, Dropout, etc.). This shows that our approach is able to efficiently model the evolution of temporal PDEs and achieve accurate uncertainty estimation,  improving the performance and trustworthiness in complex scientific and engineering problems.
\section{Related Work}
In recent years, significant efforts have been devoted to addressing the aforementioned challenges. Much of the prior work has revolved around the Bayesian formalism \cite{bernardo2009bayesian}, wherein a prior distribution is assigned to the parameters of neural networks. Subsequently, given the training data, posterior distributions over the parameters are computed to quantify predictive uncertainty. However, precise Bayesian inference poses computational challenges for neural networks, leading to the development of various approximation methods, including Laplace approximation \cite{mackay1992bayesian}, Markov chain Monte Carlo (MCMC) methods \cite{neal2012bayesian}, as well as variational Bayesian methods \cite{blundell2015weight,graves2011practical,louizos2016structured}, among others. The quality of predictive uncertainty obtained from Bayesian neural networks primarily depends on (1) the level of approximation due to computational constraints, and (2) the correctness of the chosen prior distribution, as convenient priors can result in unreasonable predictive uncertainties \cite{rasmussen2005healing}. In practice, Bayesian neural networks are often more challenging to implement and slower to train compared to non-Bayesian counterparts, necessitating a general-purpose solution that can offer high-quality uncertainty estimates with only minor modifications to the standard training pipeline.

Hence, \cite{gal2016dropout} proposed the use of Monte Carlo dropout (MC-dropout) during testing, utilizing dropout \cite{srivastava2014dropout} to estimate predictive uncertainty. Substantial research has also been conducted on approximate Bayesian interpretations of dropout \cite{gal2016dropout,kingma2015variational}. MC-dropout's implementation is relatively straightforward and yields favorable results, making it widely popular in practice. Dropout can also be interpreted as ensemble model combination \cite{srivastava2014dropout}, where predictions are averaged over an ensemble of neural networks (with shared parameters). The ensemble interpretation appears more reasonable, especially when dropout rates are not adjusted based on training data, as any sensible approximation to the true Bayesian posterior distribution must depend on the training data. This interpretation has spurred investigations into ensembles as an alternative solution for estimating predictive uncertainty.

Over time, the enhanced predictive performance resulting from the utilization of model ensembles has been increasingly acknowledged by researchers. Ensembles perform model combination, where multiple models are integrated to achieve a more robust model. Ensembles are expected to perform better when the true model lies outside the hypothesis class \cite{lakshminarayanan2017simple,wenzel2020hyperparameter}.
\section{Preliminaries}
LE-PDE-UQ builds upon the prior work of LE-PDE \cite{wu2022learninga}. LE-PDE-UQ shares LE-PDE's advantage of fast, accurate and scalable forward prediction and inverse optimization of PDEs, but also with notable innovations for uncertainty quantification. Below, we will provide a brief introduction to LE-PDE. 
The LE-PDE model architecture comprises four key components: \\
\mbox{}\hspace{8mm} $q$: \hspace{1mm} dynamic encoder:$z^k=q(U^k)$\\
\mbox{}\hspace{8mm} $r$: \hspace{1mm} static encoder: $z_p=r(p)$\\
\mbox{}\hspace{8mm} $g$: \hspace{1mm}latent evolution model: $z^{k+1}=g(z^k,z_p)$\\
\mbox{}\hspace{8mm} $h$: \hspace{1mm} decoder: 
$\hat{U}^{k+1}=h(z^{k+1})$

LE-PDE utilize the temporal bundling technique \cite{brandstetter2022message} to enhance the representation of sequential data. This approach involves grouping input states $U^k$ across a fixed interval $S$ of consecutive time steps. Consequently, each latent vector $\mathbf{z}_k$ encodes this bundle of states, and latent evolution predicts the next $\mathbf{z}$ for the subsequent $S$ steps. The parameter $S$, a hyperparameter, is adaptable to the specific problem, and setting $S=1$ results in no bundling. It's crucial to note that the dynamic encoder $q$ should feature a flattening operation and a MultiLayer Perception (MLP) head that transforms feature maps into a single fixed-length vector $\mathbf{z}\in R^{d_z}$. By doing so, the latent space's dimensionality doesn't increase linearly with input dimension, allowing substantial data compression and rendering long-term predictions more efficient.

In addition to enhancing forward simulations, LE-PDE can accelerate inverse optimization. This involves using backpropagation through time (BPTT) to adjust system parameters $p$ within the latent space, minimizing a predefined objective function $L_d[p]$. This is crucial in engineering, where optimizing boundary conditions or equation parameters is essential. LE-PDE encodes initial state $U_0$ and system parameters $p$ into latent vectors $z_0$ and $z_p$ using acquired latent space knowledge and the evolutionary model. Latent evolution takes place, and if needed, it returns to the input space to calculate $L_d[p]$. By computing the gradient of $L_d[p]$ with respect to $p$ and using methods like Adam optimization, an approximate optimal $p$ can be found. With the significantly smaller latent space dimension, this method reduces the complexity of inverse optimization. For more details on LE-PDE, refer to Appendix A.
\section{Our Approach LE-PDE-UQ}
In this section, we provide a detailed explanation of our LE-PDE-UQ method. We begin by presenting the complete architecture of the algorithm framework, as illustrated in Fig. \ref{fig1}. Subsequently, we introduce the learning objectives for effectively capturing long-term evolution. Finally, we describe the efficient inverse optimization approach within the latent space enabled by our method.
\subsection{Algorithm Framework}
The model architecture of LE-PDE-UQ consists of five components: (1) a dynamic encoder $q:\mathbb{U}\to \mathbb{R}^{d_z}$ that maps the input state $U^t=\{\mathbf{u}_i^t\}_{i=1}^N\in \mathbb{U}$ to a tuple of (latent-vector, latent-uncertainty-vector): $(\mathbf{z}^t,\mathbf{z}^t_\sigma)= q(U^t)\in \mathbb{R}^{d_z}$; (2) an (optional) static encoder $r:\mathbb{P}\to \mathbb{R}^{d_{zp}}$ that maps the (optional) system parameter $p\in\mathbb{P}$ to a static latent embedding $\mathbf{z}_p$=$r(p)$; (3) a decoder $h_\mu:\mathbb{R}^{d_z}\to \mathbb{U}$ that maps the latent vector $\mathbf{z}^t\in \mathbb{R}^{d_z}$ back to the input state $U^t$; (4) a latent evolution model $g:\mathbb{R}^{d_z}\times \mathbb{R}^{d_{zp}}\to \mathbb{R}^{d_z}$ that maps $\mathbf{z}^t, \mathbf{z}^t_\sigma\in \mathbb{R}^{d_z}$ at time $t$ and static latent embedding $\mathbf{z}_p\in \mathbb{R}^{d_{zp}}$ to $\mathbf{z}^{t+1},\mathbf{z}^{t+1}_\sigma\in \mathbb{R}^{d_z}$ at time $t+1$; (5) uncertainty decoder $h_\sigma:\mathbb{R}^{d_z}\to \mathbb{U}$ that maps the latent uncertainty vector $\mathbf{z}^t_\sigma\in \mathbb{R}^{d_z}$ back to the predicted uncertainty $U^t_\sigma$. Here the latent evolution model $g$ is decomposed as:
\begin{equation}
\begin{aligned}
    \label{eq:1}
    z^{t+1}=g_\mu(z^{t},z_p)\\
    z^{t+1}_\sigma=g_\sigma([z^{t},z^t_\sigma],z_p)\\
\end{aligned}
\end{equation}

Note that the latent vector $z^{t+1}$ only depends on $z^{t},z_p$, while the latent uncertainty vector depends on $z^{t},z_p$, and the latent uncertainty vector at $z^t_\sigma$ previous time step, modeling the propagation of uncertainty in latent space. 
We employ the temporal bundling trick \cite{brandstetter2022message} where each input state $U^t$ can include states over a fixed length $S$ of consecutive time steps. At inference time, LE-PDE-UQ performs autoregressive rollout in latent space $\mathbb{R}^{d_z}$:
\begin{equation}
\begin{aligned}
    \label{eq:2}
    (\hat{U}^{t+m},\hat{U}^{t+m}_\sigma)=(h,h_\sigma)\circ g\left(\cdot,r(p)\right)^{(m)} \circ q (\hat{U}^t)\\
    \equiv (h,h_\sigma)\bigg(\underbrace{g(\cdot,r(p))\circ...\circ g(\cdot,r(p))}_{\text{composing}\ m\ \text{times}}\left(q(\hat{U}^t)\right)\bigg)
\end{aligned}
\end{equation}

Compared to autoregressive rollout in input space, LE-PDE-UQ can significantly improve efficiency with a much smaller dimension of $\mathbf{z}^t\in \mathbb{R}^{d_z}$ compared to $U^t\in \mathbb{U}$. Moreover, it efficiently models the propagation of uncertainty in latent space, using the latent uncertainty vector $z_\sigma^t$. Here we do not limit the architecture for encoder, decoder and latent evolution models. Depending on the input $U^t$, the encoder $q$ and decoder $h_\mu$ can be a CNN or GNN with a (required) MLP head. We model the latent evolution model $g$ as an MLP with residual connection from input to output. 
During forward prediction, we also augment the above architecture with Deep Ensemble \cite{lakshminarayanan2017simple} for improved uncertainty quantification.

\subsection{ Learning Objective}
Given discretized inputs $\{U^t\},t=1,...K+M$, our LE-PDE-UQ model is trained with the following objective that combines negative log-likelihood in the input space, reconstruction, and long-term consistency in the latent space:
\begin{equation}
\label{eq:3}
L=\frac{1}{T}\sum_{t=1}^T (L^t_{\text{multi-step}} +  L^t_{\text{recons}} +  L^t_{\text{consistency}})\\
\end{equation}

Here $\ell$ is the loss function for individual predictions, which can typically be MSE or L2 loss. $\hat{U}^{t+m}$ is given in Eq. (\ref{eq:2}).  $L_{\text{recons}}^t$ aims to reduce reconstruction loss. $L_{\text{multi-step}}^t$ performs latent multi-step evolution given in Eq. (\ref{eq:2}) and compare with the target $U^{t+m}$ in input space, up to time horizon $M$. $\alpha_m$ are weights for each time step, which we find that $(\alpha_1,\alpha_2, ...\alpha_M)=(1,0.1,0.1,...0.1)$ works well. Besides encouraging better prediction in input space via $L_\text{multi-step}^t$, we also want a stable long-term rollout in latent space. This is because in inference time, we want to mainly perform autoregressive rollout in latent space, and decode to input space only when needed. Thus, we introduce a novel latent consistency loss $L_{\text{consistency}}^t$, which compares the $m$-step latent rollout $g\left(\cdot,r(p)\right)^{(m)}\circ q(U^t)$ with the latent target $q(U^{t+m})$ in \emph{latent} space. The denominator $|| q(U^{t+m})||_2^2$ serves as normalization to prevent the trivial solution that the latent space collapses to a single point. Taken together, the three terms encourage a more accurate and consistent long-term evolution both in latent and input space.

\subsection{Inverse Optimization}
In addition to improved efficiency for forward simulation, LE-PDE-UQ also allows more efficient solving of inverse problems, via backpropagation through time (BPTT) in latent space. Given a specified objective  $L_d[p,U^0]=\sum_{k=k_s}^{k_e} \ell(U^t)$ which is a discretized version of $L_d[\mathbf{a},\partial\mathbb{X}]$, we define the objective: 
\begin{equation}
\label{eq:4}
L_d[p,U^0]=\sum_{m=k_s}^{k_e} \ell_d(\hat{U}^m(p,U^0))
\end{equation}

For inverse problems that infer unknown parameters or initial state (so that all future state can be known), the objective $L_d$ can be an MSE between predicted future states $\hat{U}^m$ and the observed future states $U^m$. For inverse design tasks, $L_d$ can be specific design objectives such as lift-drag ratio for plane shape design.
$\hat{U}^m=\hat{U}^m(p,U^0)$ is given by Eq. (\ref{eq:4}) setting $k=0$ using our learned LE-PDE-UQ, which starts at initial state of $U^0$, encode it and $p$ into latent space, evolves the dynamics in latent space and decode to $\hat{U}^m$ as needed. The static latent embedding $\mathbf{z}_p=r(p)$ influences the latent evolution at each time step via $g(\cdot,r(p))$. The initial state $U^0$ influences the future state via the latent vectors $(\mathbf{z}^0,\mathbf{z}^0_\sigma)= q(U^0)$. To perform inverse optimization w.r.t. the high-dimensional initial state $U^0$, we optimize w.r.t. $(\mathbf{z}^0,\mathbf{z}^0_\sigma)$ first and then use the decoder $h$ to decode $\mathbf{z}^0$ to an estimated $\hat{U}^0$. This is different from LE-PDE where we optimize directly w.r.t the input variable. This is because $\hat{U}^0$ has a much higher dimension than $(\mathbf{z}^0,\mathbf{z}^0_\sigma)$, and optimizing w.r.t. $\hat{U}^0$ can lead to adversarial modes, as is also seen in \cite{zhao2022learning}. Instead, optimizing w.r.t. $(\mathbf{z}^0,\mathbf{z}^0_\sigma)$ then decode leads to more physical estimation of $U^0$. To obtain uncertainty for the inverse optimization, we employ Deep Ensemble \cite{lakshminarayanan2017simple} to obtain estimated uncertainty.
\section{Experiments}
In the experiments, our goal is to answer the following questions: (1) Can LE-PDE-UQ accurately quantify the uncertainty arising from the long-term evolution of complex systems and compete with state-of-the-art methods? (2) Which components of LE-PDE-UQ effectively enhance its uncertainty quantification capability in forward problem inference? (3) How does LE-PDE-UQ perform in quantifying uncertainty during the model's inverse optimization process? The experimental section on Forward Problems is primarily aimed at addressing questions (1) and (2), while the section on Inverse optimization is mainly focused on addressing question (3). We evaluate the models with two aspects: quality in uncertainty quantification measured by miscalibration area (MA), mean absolute calibration error (MACE), and root mean squared calibration error (RMSCE); and quality in point prediction, measured by relative L2 loss and mean absolute error (MAE) (see Appendix B for more details). The evaluation is generated using the Uncertainty Toolbox package \cite{chung2021uncertainty}.
\begin{table*}[t]
\centering
\resizebox{0.65\textwidth}{!}{%
\begin{tabular}{@{}l|ccc|cccc@{}}
\hline
           &MA   & MACE & RMSCE & L2 & MAE   \\
\hline
Bayes layer with Latent     &\underline{0.0445} & \underline{0.0440}  & \underline{0.0500}  & 0.2345  & 0.2051   \\
Bayes layer without Latent      &0.2381 & 0.2357 & 0.2665 & 0.2105  & 0.1830   \\
\hline
Dropout, L2=0      &0.1778 & 0.1760 & 0.1979 & 0.2079  & 0.1938    \\
Dropout, L2=$10^{-5}$     &0.1924 & 0.1905 & 0.2143 & 0.2092  & 0.1958  \\
Dropout, L2=$10^{-4}$     &0.2317 & 0.2294 & 0.2588 & 0.2458  & 0.2320   \\
Dropout, L2=$10^{-3}$    &0.3281 & 0.3248 & 0.3704 & 0.3534  & 0.3428    \\
\hline
NoLatent (single, with $\sigma$)     &0.1045 & 0.1035 & 0.1175 & 0.2053  & 0.1817   \\
NoLatent (ensemble, without $\sigma$)     &0.2118 & 0.2096 & 0.2355 & 0.1939  & 0.1657   \\
NoLatent (ensemble, with $\sigma$)     &0.0602 & 0.0596 & 0.0662 & 0.1939  & 0.1657   \\
\hline
Latent (single, without $\sigma$)     &- & - & - & \textbf{0.1890}  & \underline{0.1613}    \\
Latent (single, with $\sigma$)     &0.0576 & 0.0570 & 0.0649 & 0.2108  & 0.1811    \\
Latent (ensemble, without $\sigma$)    &0.1823 & 0.1805 & 0.2024 & \underline{0.1895}  & \textbf{0.1608}    \\
Latent (\textbf{ours}, ensemble, with $\sigma$)   &\textbf{0.0142} & \textbf{0.0141} & \textbf{0.0160} & \underline{0.1895}  & \textbf{0.1608}  \\
\hline
\end{tabular}}
\caption{
Accuracy of different methods for uncertainty quantification in forward problems. The Bayes layer with Latent method refers to its combination of the concepts of Bayesian layer and latent space. Dropout, $L2 = 10^{-5}$ indicates that this approach simultaneously utilizes both Dropout technique and L2 regularization. Under NoLatent and Latent models, `single' refers to using a single model for prediction or analysis, while ``ensemble'' refers to using an ensemble of 10 models for prediction or analysis. $\sigma$ refers to that a single model can also predict uncertainty. Bold font represents the best results among the methods, while underline  indicates second-best.
}
\label{tab:1}
\label{table1}
\end{table*}
\subsection{Dataset}
We have tested the LE-PDE-UQ within a 2D benchmark based on the Navier-Stokes equation. The Navier-Stokes equation has wide applications in science and engineering, including fields like weather forecasting and jet engine design. Simulation becomes more challenging when entering the turbulent phase, which exhibits multiscale dynamics and chaotic behavior. Specifically, we test our model in a viscous, incompressible fluid in vorticity form in a unit torus:
\begin{equation}
\label{eq:5}
\begin{split}
\partial_t w(t,x)+u(t,x)\cdot \nabla w(t,x) &= \nu \Delta w(t,x)+f(x) \\
\nabla\cdot u(t,x)&=0\\
w(0,x)&=w_0(x)\\ 
x\in(0,1)^2&,t\in[0,T]
\end{split}
\end{equation}
where $w(t,x)=\nabla\times u(t,x)$ is the vorticity, $\nu\in \mathbb{R}_+$ is the viscosity coefficient. The domain is discretized into $64\times 64$ grid and $Re=10^{4}$ (turbulent). The dataset comprises a total of 1200 trajectories (among them, 1000 trajectories are used as the training set, and an additional 200 trajectories are used as the test set), with a total of 20 time points sampled along each trajectory.
\subsection{Forward Problems}
In this section, we address questions (1) and (2). We compare the most widely used uncertainty quantification algorithms, including Bayes layer \cite{tran2019bayesian}, Dropout \cite{srivastava2014dropout} and Deep Ensembles \cite{lakshminarayanan2017simple}. We also  explore the effects of important components of our model, including latent evolution (with ablation model of NoLatent that evolves the state in input space), evolving latent uncertainty vector $\mathbf{z}_\sigma^t$, and ensembling. To ensure a fair comparison, all models utilize the past 10 steps to predict the next step, and autoregressively predict future 10 steps. 

The experiment results are shown in Table \ref{tab:1}. We see that: (1) Our full LE-PDE-UQ method attains the best performance in UQ and prediction error. (2) As demonstrated by our model (Latent) and NoLatent counterparts, latent evolution significantly improves UQ: it can propagate uncertainty via long-term rollout, (3) Comparing with and without single model uncertainty prediction (i.e., with or without $\sigma$), we see that without $\sigma$, even with ensembling, the UQ is significantly worse. Note that ``NoLatent (ensemble, with $\sigma$)'' denotes the Deep Ensembles method. (4) Bayes layer performs well in UQ, but significantly worse in prediction error. Dropout performs poorly in both UQ and prediction error.

\begin{figure}[t]
\centering
\includegraphics[width=0.40\textwidth]{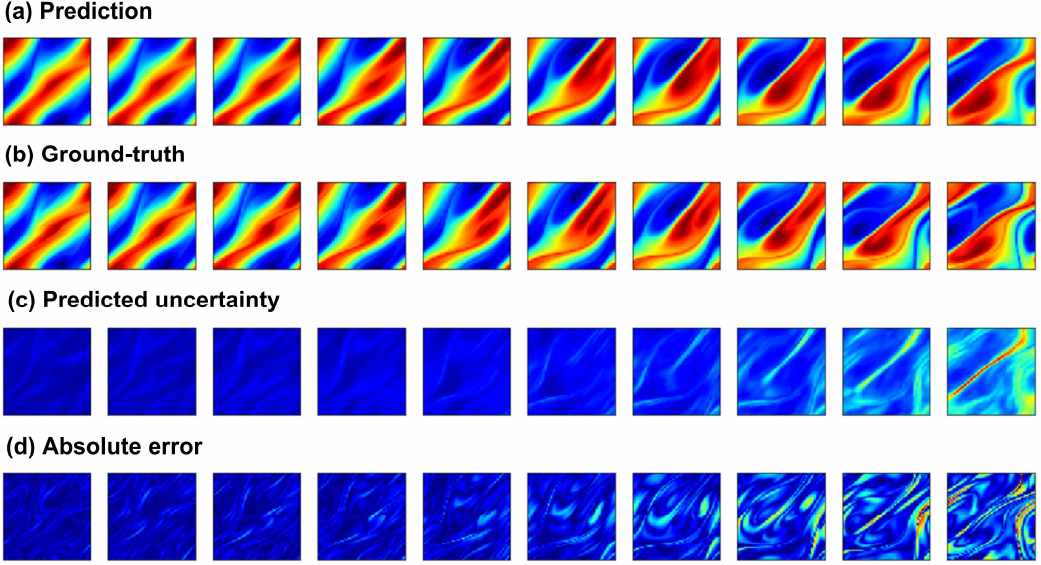}
\caption{Visualization of forward prediction results of LE-PDE-UQ on 2D Navier-Stokes turbulent flow dataset. The figure predicts the fluid state from 11-20 steps using the actual fluid state data from 1-10 steps. In this context, (a) represents the actual fluid state from 11-20 steps, while (b) indicates the fluid state predicted by LE-PDE-UQ. (c) represents the absolute error between (a) and (b), and (d) represents the uncertainty quantification results obtained by LE-PDE-UQ.} 
\label{fig2}
\end{figure}

\begin{figure}[!]
\centering
\includegraphics[width=0.45\textwidth]{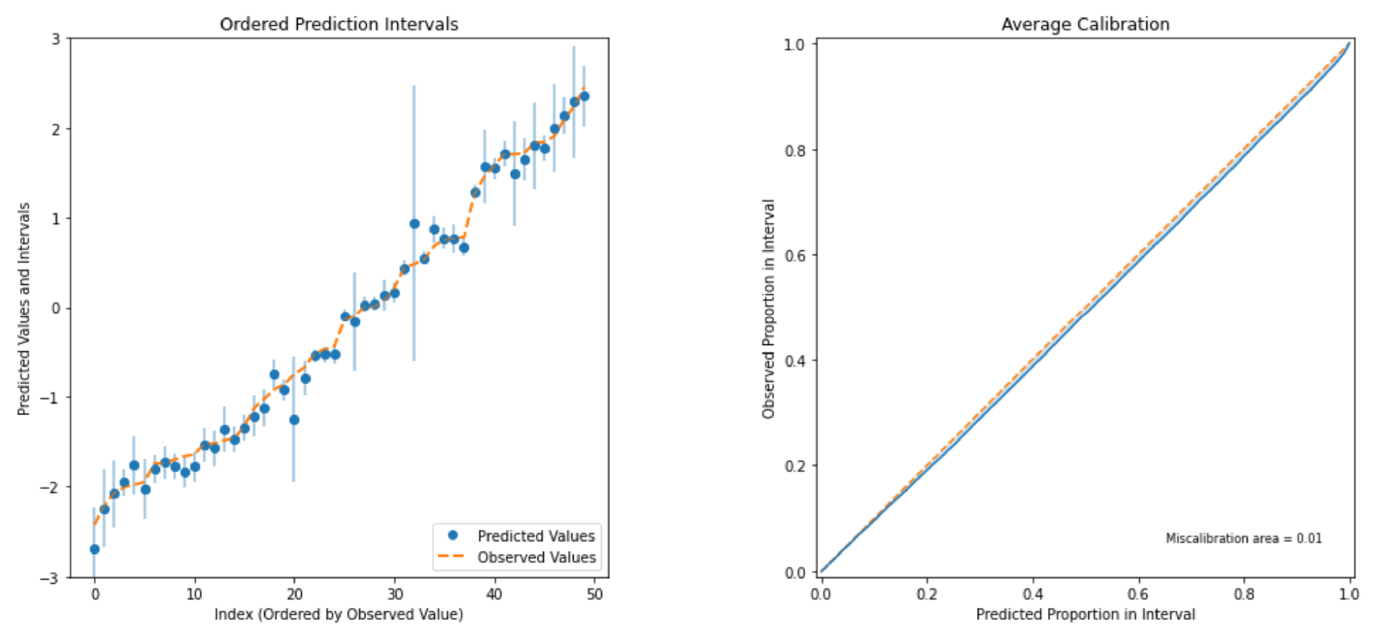}
\caption{The ordered prediction intervals and average calibration of LE-PDE-UQ.} 
\label{fig3}
\end{figure}

\begin{figure}[t]
\centering
\includegraphics[width=0.45\textwidth]{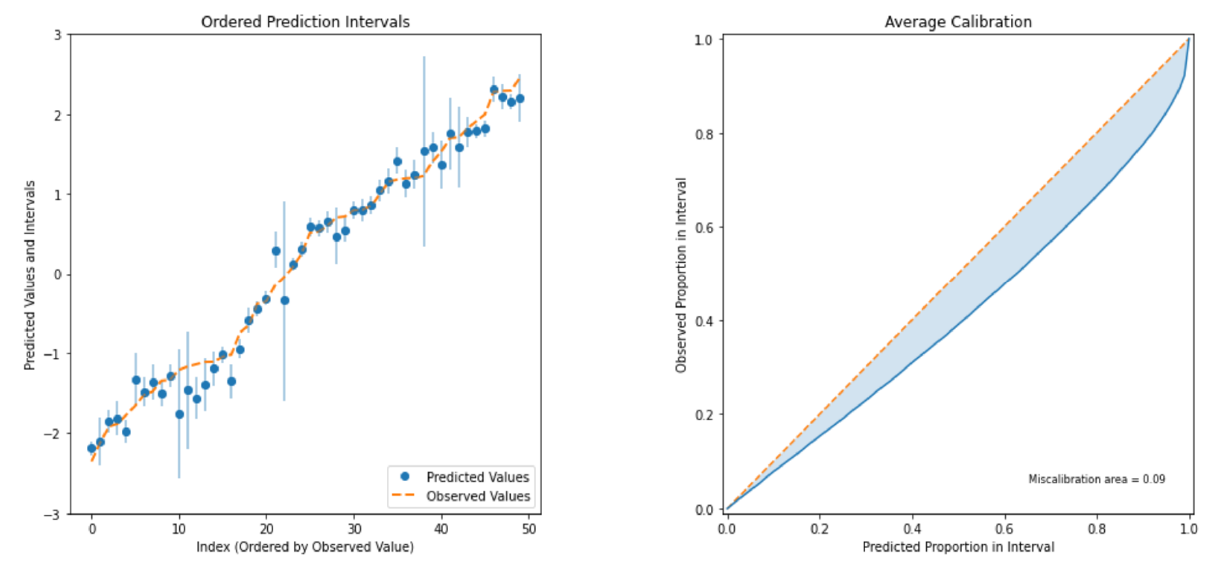}
\caption{The ordered prediction intervals and average calibration of NoLatent (ensemble, with $\sigma$).} 
\label{fig4}
\end{figure}

Fig. \ref{fig2} shows the visualization of the prediction and uncertainty quantification by our algorithm. We see that our model's prediction (first row) matches excellently with the ground-truth (second row), including both global and fine-grained spatial features. More importantly, our model's predicted uncertainty (third row) shows excellent similarity with the actual absolute error (fourth row), demonstrating accurate uncertainty quantification achieved by LE-PDE-UQ. Fig. \ref{fig3} shows the ordered prediction intervals and average calibration plot of the LE-PDE-UQ algorithm. From the left plot, it can be observed that the actual observed points (in yellow) align closely with the blue predicted region (with blue dots representing the center points of the prediction intervals). Simultaneously, in the right plot, the predicted confidence levels align precisely with the frequency of actual observations (evident from the close fit of the blue prediction line to the diagonal line), demonstrating our model's well-calibrated uncertainty quantification. We observe that the fit of the blue curve to the diagonal in the right panel of Fig. \ref{fig4} is much worse than in Fig. \ref{fig3}. From this we can conclude that Latent is critical to the ability to quantify model uncertainty.
\begin{table*}[!]
\centering
\resizebox{0.8\textwidth}{!}{%
\begin{tabular}{@{}l|ccc|cccc@{}}
\hline
Augoregressive rollout  &MA   & MACE & RMSCE & L2 & MAE  \\
\hline
NoLatent (ensemble, with $\sigma$)     &0.0602 & 0.0596 & 0.0662 & 0.1939  & 0.1657   \\
Latent (\textbf{ours}, ensemble, with $\sigma$)   &\textbf{0.0142} & \textbf{0.0141} & \textbf{0.0160} & \textbf{0.1895}  & \textbf{0.1608}  \\
\hline
\end{tabular}}
\caption{Results of Auto-regressive Rollout Experiment}
\label{table2}
\end{table*}
\begin{table*}[!]
\centering
\resizebox{0.8\textwidth}{!}{%
\begin{tabular}{@{}l|ccc|cccc@{}}
\hline
Teacher-forcing    &MA   & MACE & RMSCE & L2 & MAE  \\
\hline
NoLatent (ensemble, with $\sigma$)     &0.0260 & 0.0258 & 0.0289 & 0.1670  & 0.1420   \\
Latent (\textbf{ours}, ensemble, with $\sigma$)   &\textbf{0.0101} & \textbf{0.0100} & \textbf{0.0124} & \textbf{0.1562}  & \textbf{0.1296}  \\
\hline
\end{tabular}}
\caption{Results of Teacher-Forcing Experiment}
\label{table3}
\end{table*}
\subsubsection{Effect of latent uncertainty propagation using $\mathbf{z}_\sigma^t$.} To further investigate how much of the performance difference is attributed to the ability to model uncertainty propagation rather than model architecture differences, we conducted comparative experiments involving two distinct strategies (Autoregressive rollout and Teacher-Forcing) used by the model during inference. While Autoregressive rollout uses model's prediction as input to the next-step's prediction (used in Table \ref{table1}), Teacher-forcing provides the ground-truth as input for each step, eliminating the propagation of uncertainty.

The final experimental results are presented in Table \ref{table2} and Table \ref{table3}. In Table \ref{table2}, we see that our model (Latent)'s miscalibration area (MA) of 0.0142 is significantly smaller than Nolatent (MA=0.0602) which does not have uncertainty propagation. There are two possible causes of this gap: uncertainty propagation enabled by our model and the slight difference between the two model architectures. If we perform Teacher-forcing (Table \ref{table3}) which eliminates the effect of uncertainty propagation, the NoLatent MA reduces significantly to 0.0260. This means that the gap in Table \ref{table2} between Nolatent (MA=0.0602) and our model (MA=0.0142) is mostly due to the uncertainty propagation enabled by our model. With the latent evolution model $g$ evolving both the latent vector and latent uncertainty vector $\mathbf{z}_\sigma^t$, our model is able to accurately account for the propagation of uncertainty.
\subsubsection{Key Factors Influence.} 
\begin{figure}[t]
\centering
\includegraphics[width=0.45\textwidth]{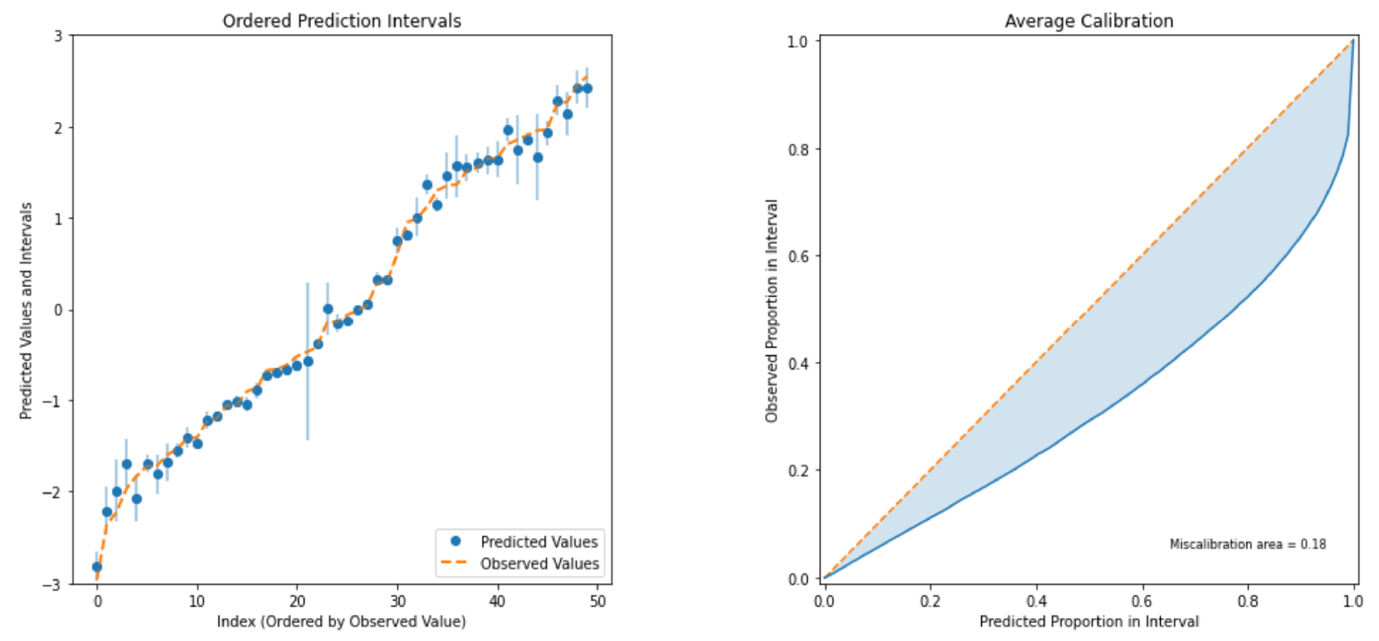}
\caption{The ordered Prediction Intervals and Average Calibration Plot Analysis for Latent full + Deterministic case.} 
\label{fig5}
\end{figure}

\begin{figure}[!]
\centering
\includegraphics[width=0.45\textwidth]{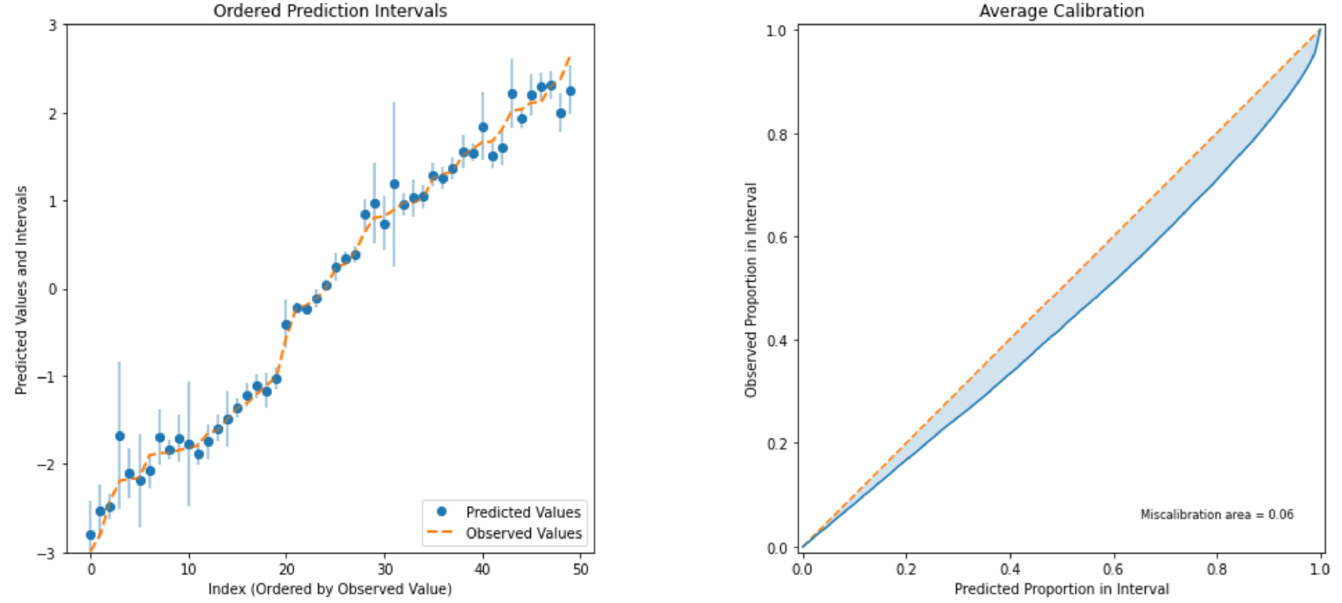}
\caption{The ordered Prediction Intervals and Average Calibration Plot Analysis for Latent full + L1 case.} 
\label{fig6}
\end{figure}

\begin{figure}[!]
\centering
\includegraphics[width=0.45\textwidth]{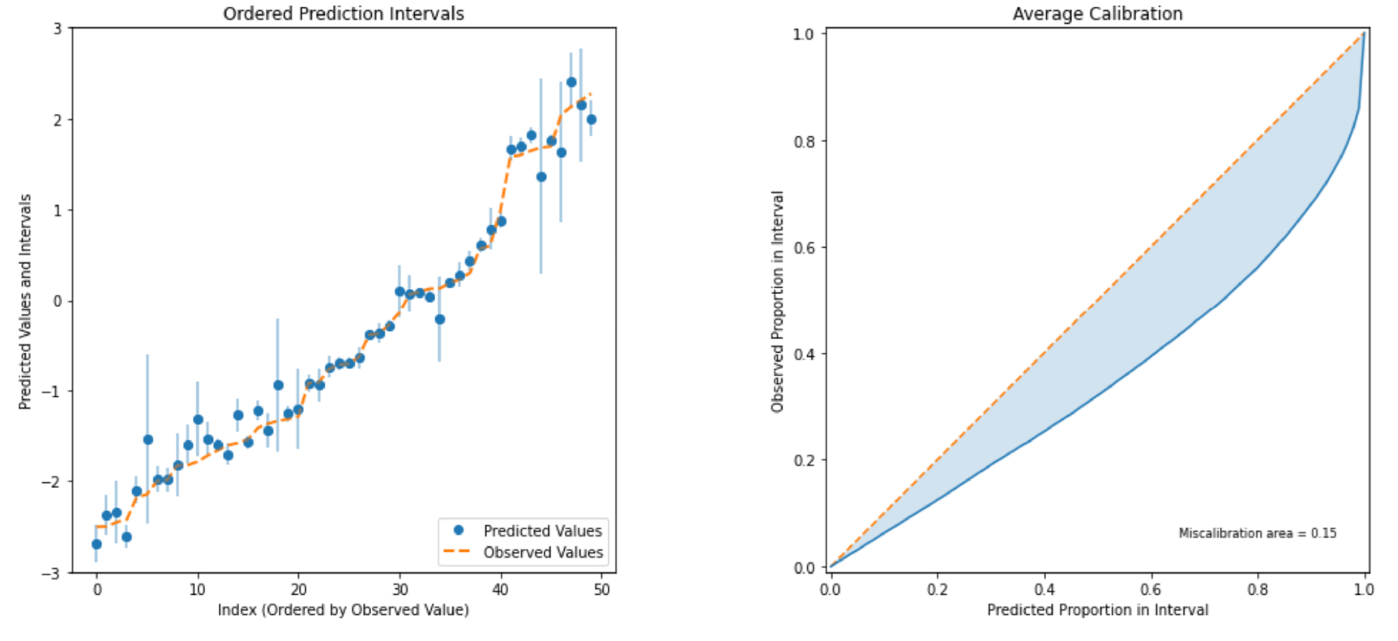}
\caption{The ordered Prediction Intervals and Average Calibration Plot Analysis for Latent full + No $\mathbf{z}_\sigma^t$ case.} 
\label{fig7}
\end{figure}

In this experiment, we primarily investigate the key factors: Deterministic, L1, and $\mathbf{z}_\sigma^t$, and their respective impacts on the uncertainty quantification performance of the latent evolution framework (Latent full) used in this paper. The specific experimental results are shown in Fig. \ref{fig5}, Fig. \ref{fig6} and Fig. \ref{fig7}. We compared these three graphs with Fig. \ref{fig3}, but due to the subtle changes in the left graph, we primarily focused on observing the right graph. We see that the effects of Deterministic and $\mathbf{z}_\sigma^t$ on the algorithm in this study are more significant than L1. We have also presented the fluid simulation images corresponding to these three experiments in Appendix C.
\begin{table*}[!]
\centering
\resizebox{0.8\textwidth}{!}{%
\begin{tabular}{@{}l|ccc|cccc@{}}
\hline
           &MA   & MACE & RMSCE & L2 & MAE  \\
\hline
NoLatent (ensemble, without $\sigma$)     &0.0929 & 0.0920 & 0.1055 & 1.5255  & 1.2580 \\
Latent (\textbf{ours}, ensemble,with $\sigma$)   &\textbf{0.0224} & \textbf{0.0222} & \textbf{0.0264} & \textbf{0.1863}  & \textbf{0.1505} \\
\hline
\end{tabular}}
\caption{Results of Inverse Optimization}
\label{table4}
\end{table*}
\subsection{Inverse Optimization}
\begin{figure}[t]
\centering
\includegraphics[width=0.40\textwidth]{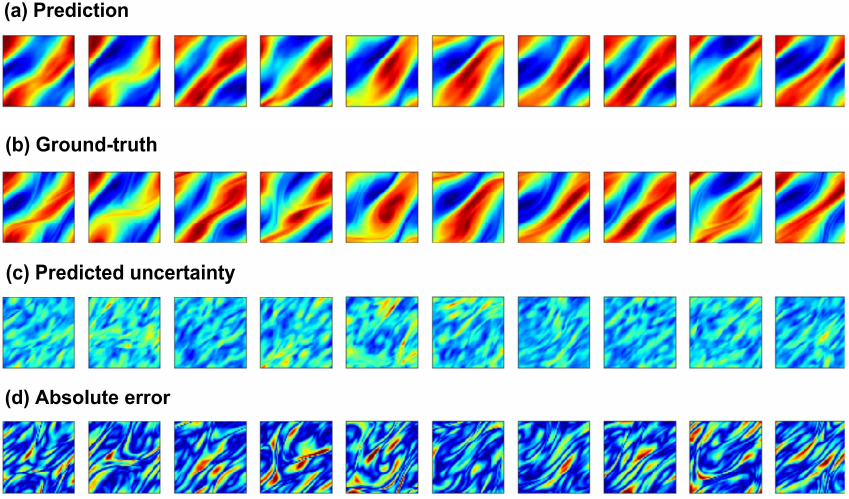}
\caption{Visualization of LE-PDE-UQ predicting the dynamics of turbulent 2D Navier-Stokes in inverse optimization.} 
\label{fig8}
\end{figure}

\begin{figure}[t]
\centering
\includegraphics[width=0.40\textwidth]{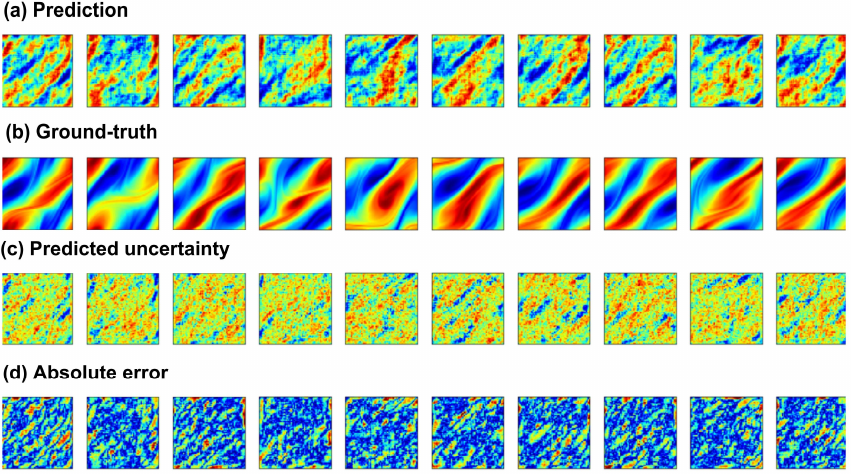}
\caption{Visualization of NoLatent (ensemble, with $\sigma$) predicting the dynamics of turbulent 2D Navier-Stokes in inverse optimization.} 
\label{fig9}
\end{figure}

\begin{figure}[t]
\centering
\includegraphics[width=0.45\textwidth]{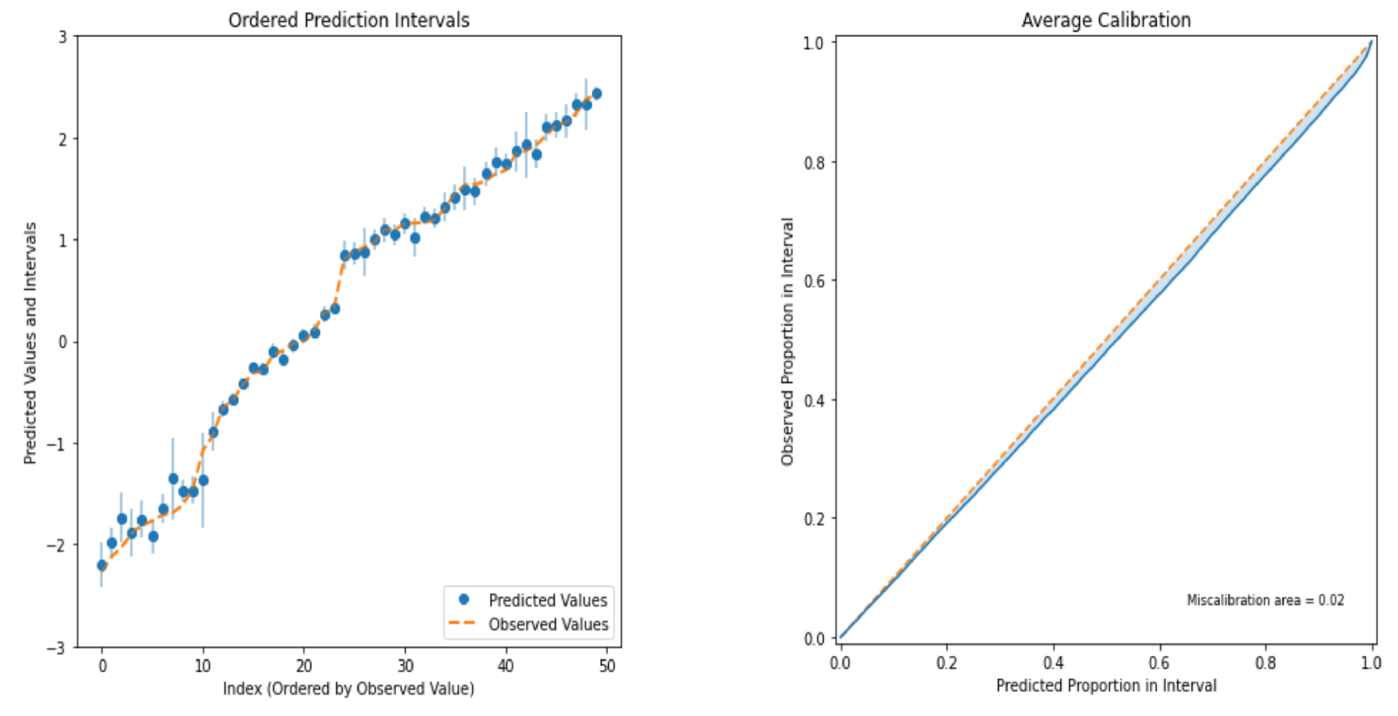}
\caption{The ordered Prediction Intervals and Average Calibration Plot Analysis for the LE-PDE-UQ in inverse optimization.} 
\label{fig10}
\end{figure}

\begin{figure}[t]
\centering
\includegraphics[width=0.45\textwidth]{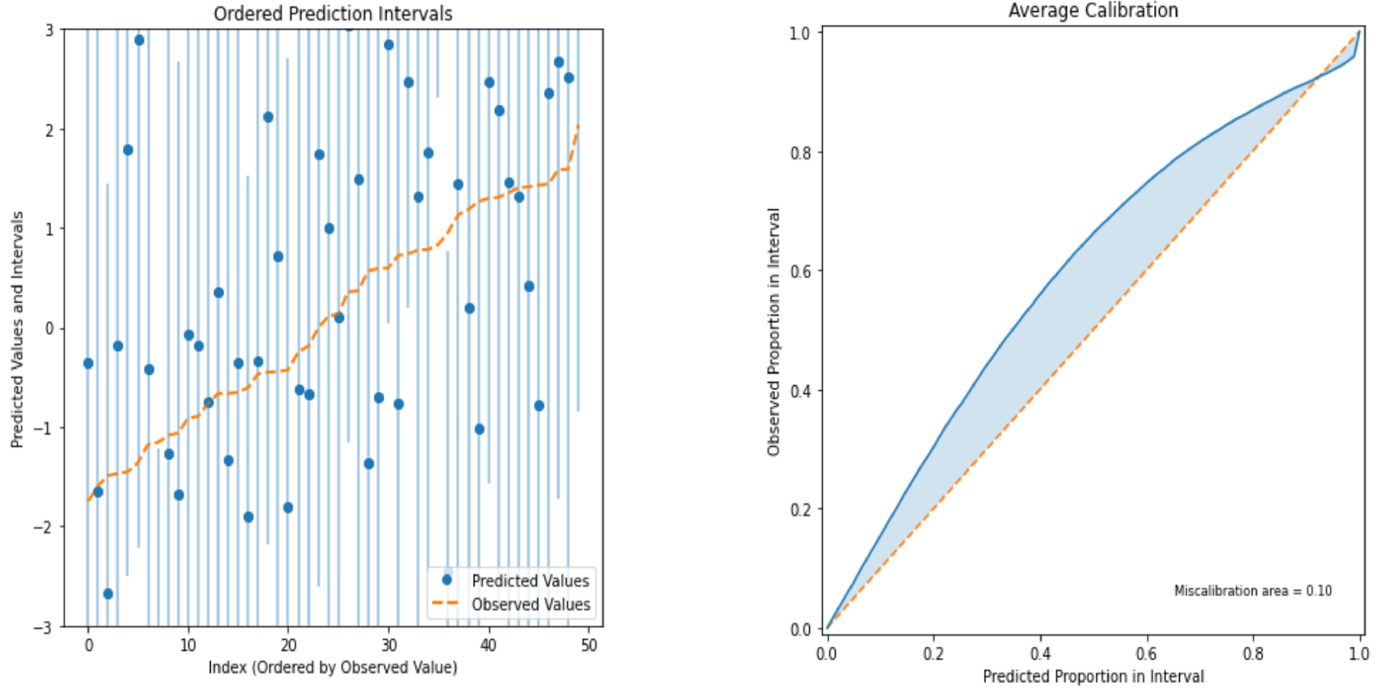}
\caption{The ordered Prediction Intervals and Average Calibration Plot Analysis for the NoLatent (ensemble, with $\sigma$) case in inverse optimization.} 
\label{fig11}
\end{figure}

In this section, our main goal is to investigate question (3) through a comparison between Latent and NoLatent approaches, and the final results are shown in Table \ref{table4}. The result shows that Nolatent has significantly larger error, and larger miscalibration error. Meanwhile, Fig. \ref{fig8} and Fig. \ref{fig10} respectively illustrate the inverse optimization  capabilities of the Latent approach (our proposed algorithm), while Fig. \ref{fig9} and Fig. \ref{fig11} depict the inverse optimization abilities of the NoLatent approach. Comparing Fig. \ref{fig8} vs. Fig. \ref{fig9} and \ref{fig10} with Fig. \ref{fig11}, we can intuitively observe that our proposed algorithm's inverse optimization predictions significantly outperform NoLatent. This is because optimizing w.r.t. the high-dimensional input space can easily find non-physical, adversarial models. In contrast, our model optimizes w.r.t. the much smaller latent dimension and then decodes back to the reasonable input space, thus achieving a much better error and uncertainty quantification.
\section{Conclusion}
In this work, we have introduced the LE-PDE-UQ framework to address the challenge of uncertainty quantification in time-dependent partial differential equations within deep learning-based surrogate models. Our method is driven by latent vectors within a dedicated latent space, enhancing the solving capabilities of both forward evolution and inverse optimization through accurate predictions and robust uncertainty estimates. Notably, LE-PDE-UQ can propagate uncertainty over extended auto-regressive rollouts without requiring additional sampling, providing a unique advantage in long-term predictions. Through rigorous experiments, our approach outperformed prominent baselines in uncertainty quantification. It demonstrated exceptional execution precision and stability in both forward and inverse scenarios, while bridging the gap between deep learning-based surrogate models and trustworthy uncertainty quantification.
\section{Acknowledgements}
We would like to express our sincere thanks to Wenhao Deng, Xianfeng Wu, Jiaxin Li and Long Wei in Westlake University AI for Scientific Simulation and Discovery Lab for discussions and for providing feedback on our manuscript. We also gratefully acknowledege the support of Westlake University Research Center for Industries of the Future and Westlake University Center for High-performance Computing. Jure Leskovec acknowledges the support of
DARPA under Nos. N660011924033 (MCS);
NSF under Nos. OAC-1835598 (CINES), CCF-1918940 (Expeditions), DMS-2327709 (IHBEM);
Stanford Data Applications Initiative,
Wu Tsai Neurosciences Institute,
Chan Zuckerberg Initiative,
Amazon, Genentech, GSK, Hitachi, Juniper Networks, KDDI, and UCB.

The content is solely the responsibility of the authors and does not necessarily represent the official views of the funding entities.
\bibliography{References}
\newpage
\hbox{}\newpage
\section{Appendix}
\makeatletter 
\renewcommand{\thefigure}{S\arabic{figure}}
\makeatother
\setcounter{figure}{0}
\setcounter{equation}{0}
\subsection{A. A detailed introduction to LE-PDE}
The states of LE-PDE in input space are $U^t$, $U^{t+1}$, ... To learn an evolution, this work evolves it in latent space using the
latent representation $z^t$. The architecture consists of the following components:
\subsubsection{Dynamic encoder $q$.}
The dynamic encoder $q$ is composed of a single CNN layer with a configuration of $(kernel-size, stride, padding) = (3, 1, 1)$, employing ELU activation. This is succeeded by $F_q$ convolution blocks, followed by a flatten operation and an MLP containing one layer with linear activation. The resulting output is a $d_z$-dimensional vector $\mathbf{z}^k\in \mathbb{R}^{d_z}$ at time step $k$. Each of the $F_q$ convolution blocks comprises a convolution layer with $(kernel-size, stride, padding) = (4, 2, 1)$, followed by group normalization (Wu and He 2018) with the number of groups set to 2, and ELU activation (Clevert, Unterthiner, and Hochreiter 2015). The channel size of each convolution block follows an exponential growth pattern: the first convolution block has $C$ channels, the second has $C \times 2^1$ channels, and so forth. The channel size increases exponentially to partly address the reduced spatial dimensions of the feature map in higher layers.
\subsubsection{Static encoder $r$.}
Regarding the static encoder $r$, it adapts based on the static parameter $p$, allowing for either an MLP architecture with $F_{r'}$ layers or a CNN+MLP setup similar to the dynamic encoder. In the case of an MLP, the architecture includes $F_{r'}$ layers utilizing ELU activation, with the final layer implementing linear activation. In LE-PDE, they opt for $F_{r'}\in{0,1,2}$. When $F_{r'}=0$, it signifies no layers, and the static parameter is used directly as $\mathbf{z}_p$. The outcome of the static encoder is a $d_{z_p}$-dimensional vector $\mathbf{z}_p\in \mathbb{R}^{d_{z_p}}$.
\subsubsection{Latent evolution model $g$.}
The latent evolution model $g$ receives the combined input of $\mathbf{z}^k$ and $\mathbf{z}_p$ (concatenated along the feature dimension) and generates the prediction $\hat{\mathbf{z}}^{k+1}$. In this case, LE-PDE approach models it as an MLP featuring a residual connection extending from input to output, effectively resembling the forward Euler's method in latent space:
\begin{equation}
\label{eq:S1}
\hat{\mathbf{z}}^{k+1}=\text{MLP}_g(\mathbf{z}^k)+\mathbf{z}^k
\end{equation}
\begin{figure*}[t]
\centering
\includegraphics[width=0.8\textwidth]{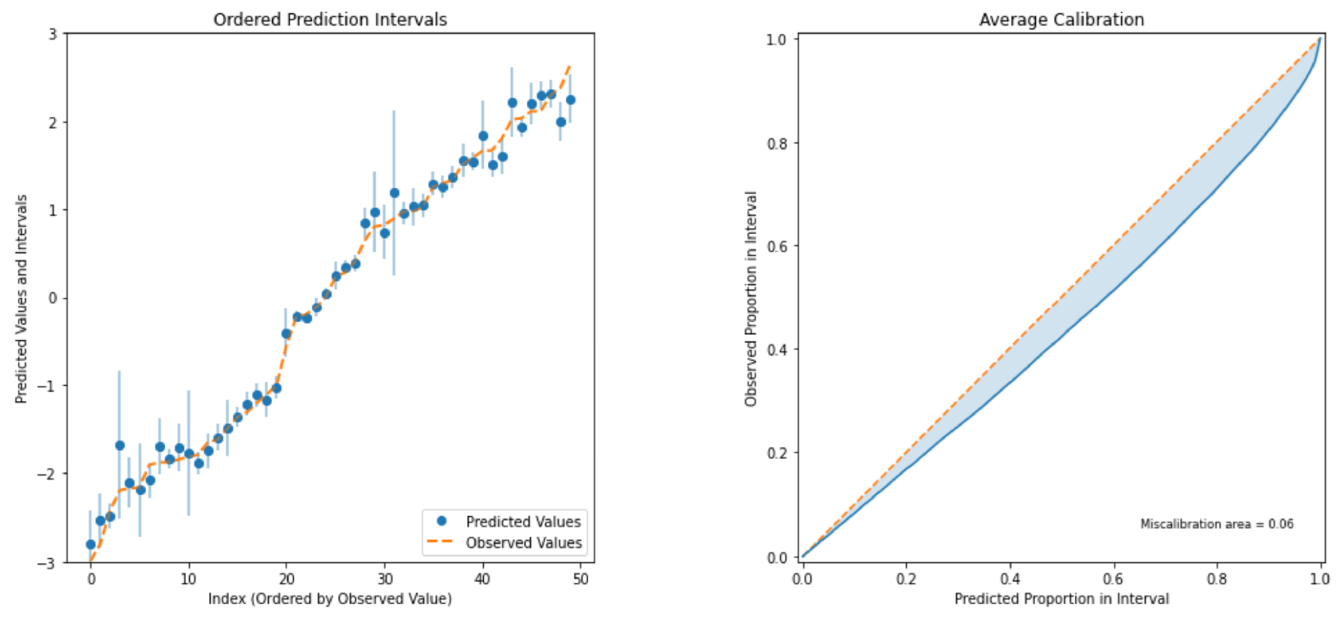}
\caption{The illustrations regarding examples of Ordered Prediction Intervals and Average Calibration.} 
\label{figS1}
\end{figure*}

In all sections of LE-PDE, the consistent architecture of $\text{MLP}_g$ is employed. This $\text{MLP}_g$ structure encompasses a total of 5 layers, with each layer containing the same number of neurons, $d_z$, as the dimension of $\mathbf{z}^k$. Activation follows ELU for the initial three layers, while the last two layers implement linear activation. Notably, the approach employs two linear layers in lieu of one, incorporating implicit rank-minimizing regularization (Jing, Zbontar et al. 2020). This adaptation has been found to yield superior performance compared to utilizing a single last linear layer.
\subsubsection{Decoder $h$.}
Similarly to the encoding process ($q$), the decoding process ($h$) takes $\mathbf{z}^{k+m}\in\mathbb{R}^{d_z},m=0,1,...M$ as input. This input is passed through an $\text{MLP}_h$ and a CNN structure comprising $F_h=F_q$ convolution-transpose blocks. The outcome is a mapping to the state $U^{k+m}$ within the input space. The $\text{MLP}_h$ is implemented as a one-layer MLP with linear activation. Subsequently, the vector is reshaped into the format of (batch-size, channel-size, *image-shape) to accommodate the subsequent $F_h$ convolution-transpose blocks. This transformation is followed by a single convolution-transpose layer with (kernel-size, stride, padding)=$(3,1,1)$ and linear activation. Each convolution-transpose block includes a convolution-transpose layer with (kernel-size, stride, padding)=$(4,2,1)$, followed by group normalization and ELU activation. The channel count in each block mirrors the pattern established in encoder $q$, where proximity to the output results in smaller channel sizes exhibiting exponential reduction.

To perform autoregressive rollout:
\begin{equation}
\begin{aligned}
    \label{eq:S2}
    \hat{U}^{t+m}&=h(z^{t+m})\\
    &\equiv h(g(\cdot,z_p)^{(m)}\circ z^t)\\
    &\equiv h\bigg(g(\cdot,r(p))^{(m)}\circ q({U^t})\bigg)\\
\end{aligned}
\end{equation}

The training objective is given by:
\begin{equation}
\label{eq:S3}
L=\frac{1}{T}\sum_{t=1}^T (L^t_{\text{multi-step}} +  L^t_{\text{recons}} +  L^t_{\text{consistency}}).\\
\end{equation}
\begin{equation}
\label{eq:S4}
\hspace{-50pt}\scalebox{0.9}{$\left\{
\begin{array}{ll}
L_{\text{multi-step}}^t=\sum_{m=1}^M \alpha_m\ell(\hat{U}^{t+m}, U^{t+m})\\
L_{\text{recons}}^t=\ell(h( q(U^t)),U^t)\\
L_{\text{consistency}}^t=\sum_{m=1}^M\frac{||(g(\cdot,r(p))^{(m)}\circ q(U^t))-q(U^{t+m})||_2^2}{|| q(U^{t+m})||_2^2}
\end{array}
\right.$}
\end{equation}
Here, $\ell$ represents the loss function for individual predictions, which typically comprises MSE or L2 loss. $\hat{U}^{t+m}$ is detailed in Eq. (\ref{eq:S2}). The purpose of $L_{\text{recons}}^t$ is to minimize reconstruction loss. Meanwhile, $L_{\text{multi-step}}^t$ entails latent multi-step evolution, as specified in Eq. (\ref{eq:S2}), and subsequently compares the outcomes with the target $U^{t+m}$ within the \emph{input} space, up to the time horizon $M$. The weight factors $\alpha_m$ correspond to each time step, and observations show that $(\alpha_1,\alpha_2, ...\alpha_M)=(1,0.1,0.1,...0.1)$ yields satisfactory results. Beyond fostering improved predictions in the input space through $L_\text{multi-step}^t$, a stable long-term rollout within the latent space is also sought. This stability is vital as, during inference, the objective is primarily to execute autoregressive rollout in the latent space, and to translate to the input space only when required. Thus, a fresh latent consistency loss $L_{\text{consistency}}^t$ is introduced. This loss contrasts the $m$-step latent rollout $g\left(\cdot,r(p)\right)^{(m)}\circ q(U^t)$ with the latent target $q(U^{t+m})$ within the \emph{latent} space. The denominator $|| q(U^{t+m})||_2^2$ is employed for normalization, averting the trivial scenario where the latent space collapses to a single point. Taken together, these three terms collectively promote enhanced and congruent long-term evolution in both latent and input spaces.
\subsection{B. Evaluation Metrics}
Many works in Uncertainty Quantification (UQ) often exhibit inconsistency in the evaluation metrics used, signaling a divergence on whether certain metrics should or should not be employed. For instance, some studies report likelihood ratios on test sets (Lakshminarayanan, Pritzel, and Blundell 2017; Skafte, Jørgensen, and Hauberg 2019; Zhao, Ma, and Ermon 2020), others utilize alternative appropriate scoring rules (Maciejowska, Nowotarski, and Weron 2016; Askanazi et al. 2018), while some focus on calibration metrics (Kuleshov, Fenner, and Ermon 2018; Cui, Hu, and Zhu 2020). Furthermore, due to variations in the implementation of each metric, even if similar metrics are employed, numerical results reported across different papers' contexts often cannot be directly compared. Therefore, to facilitate subsequent experimental comparisons, we employ the uncertainty toolbox package (Chung et al. 2021) in Python to compare the quantification of uncertainty achieved by the LE-PDE-UQ framework in our study. Additionally, we utilized the visualizations of the ordered prediction intervals and average calibration plot from this toolkit.
\subsubsection{Ordered Prediction Intervals.}
The Ordered Prediction Intervals graph is used to display the coverage and width of prediction intervals. Prediction intervals refer to the range of predicted values for unknown data points, representing the model's uncertainty. Coverage indicates the proportion of true values within the prediction interval, while width represents the size of the prediction interval. The graph's horizontal axis represents the index of data points, and the vertical axis represents predicted values or true values. Data points are sorted by the width of the prediction interval, from small to large, to observe the coverage of prediction intervals with different widths. Blue line segments represent prediction intervals, and orange dots represent true values. Orange dots inside the blue line segments indicate that the prediction interval covers the true value, while dots outside indicate non-coverage.

From the graph, we can understand the degree of uncertainty that the model has in predicting different data points and whether the prediction intervals accurately encompass true values. For example, in Fig. \ref{figS1} on the left, the blue intervals at the horizontal coordinate of 50 do not include the orange dots, the same phenomenon can also be observed in Fig. \ref{fig4} of the main text. This indicates that the prediction intervals do not accurately cover the true values. In other words, the model's estimation of uncertainty for these data points is not sufficiently accurate. This difference suggests that the model's predicted range does not accurately include true values with the expected confidence level, which may imply calibration or prediction accuracy issues in the model.
\subsubsection{Average Calibration.}
The Average Calibration plot is used to showcase the calibration of the model. Calibration refers to whether the model's assessment of its own prediction uncertainty aligns with reality. For instance, if the model gives a confidence of 0.9, it should be accurate with a probability of 0.9. In this plot, the x-axis represents the confidence level, while the y-axis represents the observed interval proportion (empirical coverage). Empirical coverage indicates the proportion of cases where the predicted interval contains the true value at a given confidence level. Ideally, empirical coverage should match the confidence level, which means the better the model's uncertainty prediction, the blue curve would align along the diagonal (orange) line. This indicates that the model's self-assessment of uncertainty is accurate. If empirical coverage exceeds the confidence level, it means the model lacks confidence and provides overly large prediction intervals. Conversely, if empirical coverage falls below the confidence level, it signifies the model is overly confident (as shown in the right side of Fig. \ref{figS1}) and provides overly narrow prediction intervals, the same applies to Fig. \ref{fig4} in the main text. From this plot, we can determine whether the model is well-calibrated and identify which confidence levels require calibration.
\subsubsection{How to observe and compare fluid visualization.}
\begin{figure}[!]
\centering
\includegraphics[width=0.5\textwidth]{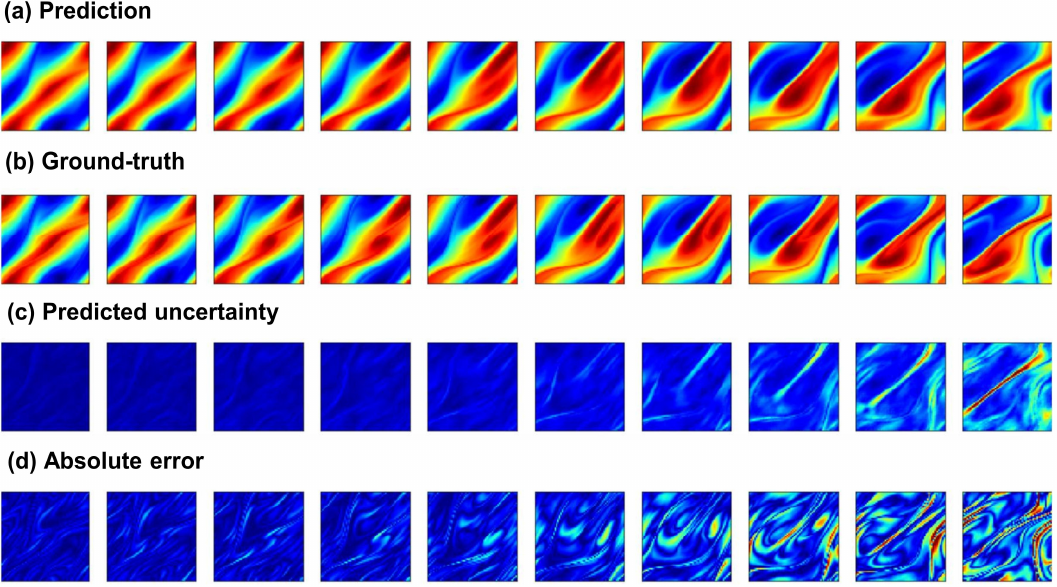}
\caption{Visualization for  the Latent full + Deterministic case testing on predicting
the dynamics of turbulent 2D Navie in forward problem.} 
\label{figS2}
\end{figure}

\begin{figure}[!]
\centering
\includegraphics[width=0.5\textwidth]{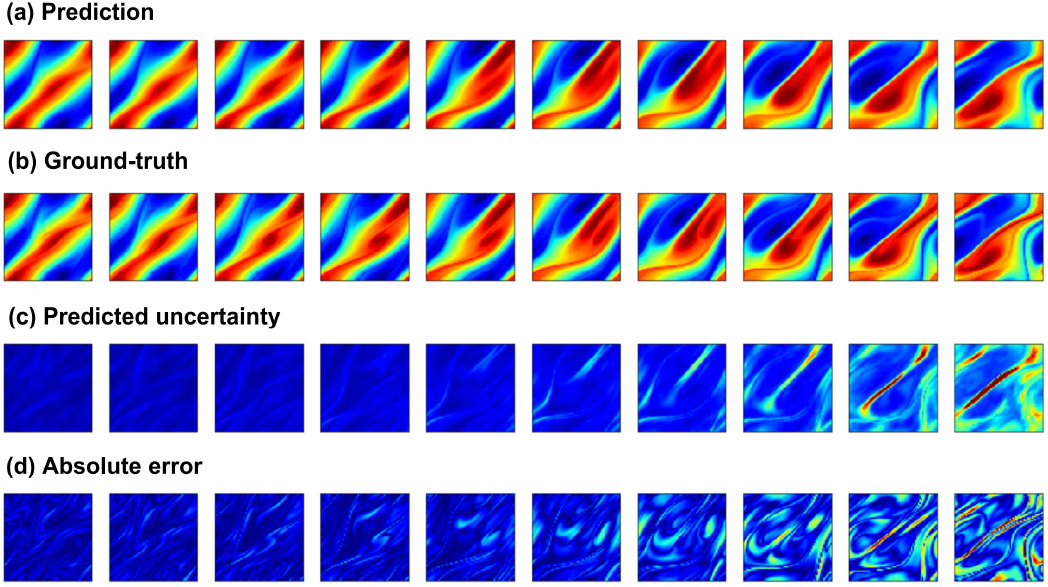}
\caption{Visualization for  the Latent full + L1 case testing on predicting
the dynamics of turbulent 2D Navie in forward problem.} 
\label{figS3}
\end{figure}

\begin{figure}[!]
\centering
\includegraphics[width=0.5\textwidth]{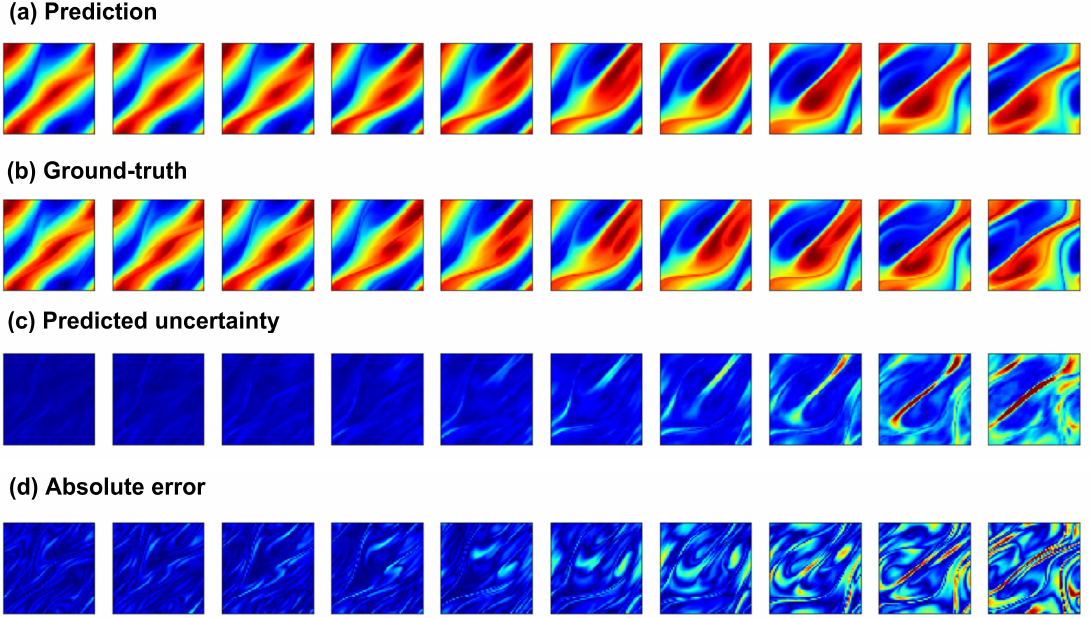}
\caption{Visualization for  the Latent full + no $\mathbf{z}_\sigma^t$ case testing on predicting
the dynamics of turbulent 2D Navie in forward problem.} 
\label{figS4}
\end{figure}

Regarding the fluid schematic diagrams (including Fig. \ref{fig2}, Fig. \ref{fig8} and Fig. \ref{fig9} in the main text, as well as Appendix Fig. \ref{figS2}, Fig. \ref{figS3} and Fig. \ref{figS4}), the way to observe and compare them can be explained as follows: The "Predicted uncertainty" indicates the model's estimation of confidence or uncertainty regarding the predicted values. On the other hand, the "Absolute error" represents the absolute difference between the model's predicted values ((a) Prediction) and the actual values ((b) Ground-truth), serving as a measure of prediction accuracy. In general, when the model exhibits higher uncertainty in its predictions, the absolute error in predictions might also be relatively larger. This is because increased uncertainty implies the model itself is less confident in its predictions, thus potentially leading to decreased accuracy. Conversely, when the model's uncertainty is lower, the absolute error in predictions could be comparatively smaller.
If the "predicted uncertainty (c) and absolute error (d) in the fluid schematic diagram appear more similar, it indicates that the model's ability to quantify uncertainty is better. This correspondence suggests that the model's uncertainty estimation aligns well with the actual error distribution.
\subsubsection{MA (Miscalibration area).}The Miscalibration area is a metric used to assess probability calibration, measuring the disparity between the probability distribution predicted by a model and the actual observed distribution. A well-calibrated model should accurately predict the probabilities of events, aligning its predicted probabilities with the observed frequencies. Hence, if the differences between the model's predicted probabilities and the observed frequencies are small, the value of the Miscalibration area will be low, indicating better calibration of the model.
In fact, the goal of the Miscalibration area is to approach zero, where the model's predicted points on the reliability diagram align closely with the diagonal line (perfect calibration). This indicates that the model's uncertainty estimation is highly consistent with the actual observations. As a result, the closer the Miscalibration area value is to zero, the better the model's performance is in terms of probability calibration.
\subsubsection{MACE (Mean Absolute Confidence Error).}The average absolute difference between the predicted uncertainties (confidence estimates) and the actual uncertainties. It measures the accuracy of the uncertainty estimates.
A lower value of MACE indicates more accurate estimation of the model's uncertainty.
\subsubsection{RMSCE (Root Mean Squared Confidence Error).}RMSCE is a metric used to assess the uncertainty estimation of predictive models. It is employed to evaluate the disparity between the model's uncertainty predictions and actual observations. Unlike the Mean Absolute Confidence Error (MACE), RMSCE utilizes squared errors to measure the differences and then takes the square root. This allows larger errors to have a greater impact, emphasizing cases with more significant discrepancies. Lower RMSCE values, compared to models with smaller errors, indicate more accurate uncertainty estimation by the model.
\subsubsection{L2 (L2 Loss or Euclidean Loss).}L2 is a type of loss function used to measure the error of a predictive model. It calculates the sum of squared Euclidean distances between predicted values and true values. L2 loss is commonly employed in various machine learning and deep learning tasks, aiming to minimize the squared differences between predictions and actual values to optimize model performance. Smaller values of L2 loss indicate better performance.
\subsubsection{MAE (Mean Absolute Error).}In the context of uncertainty quantification, MAE typically refers to Mean Absolute Error. It is a common metric used to assess the quality of uncertainty estimation in predictive models.

The Mean Absolute Error measures the average absolute difference between the model's uncertainty estimation for each predicted value and the actual observations. In uncertainty quantification, this metric is often employed to gauge the disparity between the probability distribution predicted by the model and the actual observed distribution.

The specific steps to calculate the Mean Absolute Error are as follows: (1) Calculate the expected value of the predicted probability distribution for each predicted value (e.g., the mean). (2) Calculate the actual observed value for each predicted value. (3) Compute the absolute error between the prediction and the observation for each predicted value. (4) Average the absolute errors for all predicted values to obtain the Mean Absolute Error. A smaller MAE value indicates that the model's uncertainty estimation is closer to the actual observations, thereby reflecting better quality of uncertainty estimation.
\subsection{C. Visualization of Experimental Results}
The Fig. \ref{figS2}, Fig. \ref{figS3} and Fig. \ref{figS4} respectively represent the visualized results of fluid prediction and uncertainty quantification for Latent full under the conditions of Deterministic, L1, or no $\mathbf{z}_\sigma^t$. By comparing these three figures, it is evident that Fig. \ref{figS2} and Fig. \ref{figS4} exhibit significantly lower accuracy in uncertainty quantification compared to Fig. \ref{figS3}. This further supports our main conclusion that the impact of factors like Deterministic and no $\mathbf{z}_\sigma^t$ on the uncertainty quantification ability of Latent full is greater than the influence of the L1 factor.\\


\section*{References}
Lakshminarayanan, B.; Pritzel, A.; and Blundell, C. 2017. Simple and scalable predictive uncertainty estimation using deep ensembles. \textit{Advances in neural information processing systems}, 30.\\
Skafte, N.; Jørgensen, M.; and Hauberg, S. 2019. Reliable training and estimation of variance networks. \textit{Advances in Neural Information Processing Systems}, 32.\\
Zhao, S.; Ma, T.; and Ermon, S. 2020. Individual calibration
with randomized forecasting. \textit{In International Conference
on Machine Learning}, 11387–11397. PMLR.\\
Maciejowska, K.; Nowotarski, J.; and Weron, R. 2016. Probabilistic forecasting of electricity spot prices using Factor Quantile Regression Averaging. \textit{International Journal of
Forecasting}, 32(3): 957–965.\\
Askanazi, R.; Diebold, F. X.; Schorfheide, F.; and Shin, M.
2018. On the comparison of interval forecasts. \textit{Journal of
Time Series Analysis}, 39(6): 953–965.\\
Kuleshov, V.; Fenner, N.; and Ermon, S. 2018. Accurate uncertainties for deep learning using calibrated regression. In \textit{International conference on machine learning}, 2796–2804.
PMLR\\
Cui, P.; Hu, W.; and Zhu, J. 2020. Calibrated reliable regression using maximum mean discrepancy. \textit{Advances in Neural Information Processing Systems}, 33: 17164–17175.\\
Wu, Y.; and He, K. 2018. Group normalization. In Proceedings of the European conference on computer vision (ECCV), 3–19.\\
Clevert, D.-A.; Unterthiner, T.; and Hochreiter, S. 2015. Fast and accurate deep network learning by exponential linear units (elus). arXiv preprint arXiv:1511.07289.\\
Jing, L.; Zbontar, J.; et al. 2020. Implicit rank-minimizing autoencoder. Advances in Neural Information Processing Systems, 33: 14736–14746.
\end{document}